\documentclass[journal]{IEEEtran}

\usepackage{array}
\usepackage{cite}
\usepackage{amsmath,amssymb,amsfonts}
\usepackage{algorithm}
\usepackage{algorithmic}
\usepackage{textcomp}
\usepackage{multirow}
\usepackage{graphicx}
\usepackage{threeparttable}
\ifCLASSINFOpdf
\else
\fi

\begin{document}

\title{Towards Understanding Residual and Dilated Dense Neural Networks via Convolutional Sparse Coding}
%
%
%
\author{Zhiyang Zhang and Shihua Zhang*
\thanks{Zhiyang Zhang and Shihua Zhang are with the NCMIS, CEMS, RCSDS,
Academy of Mathematics and Systems Science, CAS, Beijing 100190,
and School of Mathematical Sciences, University of Chinese Academy of Sciences,
Beijing 100049, China.

*Corresponding author. Email: zsh@amss.ac.cn.}}

%
%

\markboth{ }%
{Shell \MakeLowercase{\textit{et al.}}: Bare Demo of IEEEtran.cls for IEEE Journals}
%



\maketitle

\begin{abstract}
Convolutional neural network (CNN) and its variants have led to many state-of-art results in various fields. However, a clear theoretical understanding about them is still lacking. Recently, multi-layer convolutional sparse coding (ML-CSC) has been proposed and proved to equal such simply stacked networks (plain networks). Here, we think three factors in each layer of it including the initialization, the dictionary design and the number of iterations greatly affect the performance of ML-CSC. Inspired by these considerations, we propose two novel multi-layer models--residual convolutional sparse coding model (Res-CSC) and mixed-scale dense convolutional sparse coding model (MSD-CSC), which have close relationship with the residual neural network (ResNet) and mixed-scale (dilated) dense neural network (MSDNet), respectively. Mathematically, we derive the shortcut connection in ResNet as a special case of a new forward propagation rule on ML-CSC. We find a theoretical interpretation of the dilated convolution and dense connection in MSDNet by analyzing MSD-CSC, which gives a clear mathematical understanding about them. We implement the iterative soft thresholding algorithm (ISTA) and its fast version to solve Res-CSC and MSD-CSC, which can employ the unfolding operation for further improvements. At last, extensive numerical experiments and comparison with competing methods demonstrate their effectiveness using three typical datasets.
\end{abstract}


%
\IEEEpeerreviewmaketitle

\section{Introduction}
\IEEEPARstart{N}{owadays}, neural networks have become effective techniques in many fields including computer vision, natural language processing, bioinformatics and so on. Their predecessor perceptron was proposed by Rosenblatt in 1958 \cite{Rosenblatt1958}. However, perceptron is even too simple to solve the XOR problem. To tackle more complex problems, multilayer perceptron (MLP) has been proposed. Neural networks can be seen as generalized MLPs with more special operations. The activation functions (e.g., Sigmoid, Tanh and ReLU (rectified linear unit) \cite{Jarrett2009}, \cite{Nair2010}, \cite{Glorot2011}) have been used for computing the outputs of hidden layers in neural networks, and have been considered to simulate the sparse activation of neurons in human brain.

Convolution is another important operation used for processing data that has a known, grid-like topology. For example, time-series data can be a one-dimensional grid data and image can be thought as a two-dimensional grid of pixels. Convolution operation simulates human eyes, capturing features locally and scanning globally. The range that features captured from is called the receptive field \cite{Hubel1962}, \cite{Hubel1959}. Comprehensive experiments have shown that convolution is effective to extract abstract features from data. The first CNN architecture LeNet proposed in 1998 has been successfully applied to handwritten digit recognition \cite{LeCun1998}.

Recent years, rapid improvements of hardwares and public availability of highly optimized softwares \cite{Klockner2012}, \cite{Abadi2016}, \cite{Jia2014} make it possible to train a neural network with a large number of parameters. AlexNet \cite{Krizhevsky2012} is such a classical CNN architecture, which draws our attention to deep convolutional neural networks (DCNNs). Deeper networks have strong ability to fit complex distributions, so it is easier to achieve better performance than ever before. “The deeper, the better” becomes a belief \cite{Simonyan2014}. But DCNNs are hard to train because of diverse optimization issues. The problem of vanishing or exploding gradients \cite{Bengio1994}, \cite{Glorot2010} is a notorious one, and has been addressed by two novel tricks: normalized initialization \cite{Glorot2010}, \cite{He2015} and batch normalization (BN) \cite{Ioffe2015}. When deep neural networks are trained with these two tricks, a degradation phenomenon has been exposed \cite{He2016} that deep networks achieve lower accuracy than shallow networks. ResNet \cite{He2016} is a special architecture with skip connections tackling this phenomenon. Difficulties have been settled but the optimization issues behind the degradation phenomenon solved by ResNet is still not clear.

Densely connected CNN \cite{Li2018} and mixed-scale dense convolutional neural networks (MSDNet) \cite{Pelt2018} are also well-known architectures with skip connections. Usually, they have fewer parameters to overcome the overfitting issue. Besides, there are many other architectures and methods, such as Dropout \cite{Srivastava2014}, VGG \cite{Simonyan2014}, GoogLeNet \cite{Szegedy2015}, R-CNN \cite{Girshick2014}, YOLO \cite{Redmon2016}, FCN \cite{Long2015}. All these have become very important solutions in neural networks.

However, we still do not understand the principle of CNNs clearly. All the above successes are mainly based on empirical exploration. A clear and profound theoretical understanding of such neural networks is still lacking. On the one hand, architectures with excellent performance are hard to interpret strictly. On the other hand, the design of architectures mainly depends on intuition or inspiration. The lacking of theory is currently a key problem, which limits the further development of neural networks. This situation makes us unsure when applying neural networks to some challenging fields such as self-driving, medical diagnosis and identity recognition.

Very recently, connection between convolutional sparse coding (CSC) and CNN has been established \cite{Papyan2017a}, \cite{Papyan2017b}, \cite{Sulam2019}. It brings a fresh view to CNN. In sparse coding, one assumes that a signal can be represented as a linear combination of a few columns from a matrix called dictionary, and the linear combination can be written into a sparse vector. The task of retrieving the sparsest representation of a signal over a dictionary is called sparse coding or basis-pursuit (BP) \cite{Donoho2003}, \cite{Chen1998}, \cite{Chen2001}, \cite{Tropp2006}. It is also known in the statistical learning community as the least absolute shrinkage and selection operator (Lasso) problem \cite{Tibshirani1996}. Neuroscience indicates that sparse coding or representation plays an important role in human brain \cite{Spanne2015}. Moreover, sparsity has been shown to be a driving force in a myriad of applications in computer vision \cite{Mairal2010}, \cite{Wright2010}, \cite{Zeiler2010}, statistics \cite{Tibshirani1996}, \cite{Hastie2015} and machine learning \cite{Henaff2011}, \cite{Ranzato2007}, \cite{Kavukcuoglu2010}. For a given dictionary, orthogonal matching pursuit (OMP) \cite{Pati1993}, \cite{Chen1989}, iterative soft thresholding algorithm (ISTA) and its fast version (FISTA) \cite{Beck2009} have been proposed to tackle the pursuit problem. Besides, double sparsity has also been proposed to accelerate the training process \cite{Rubinstein2009}, and it assumes the dictionary in sparse coding can be factorized into a multiplication of two matrices.

Inspired by these progresses, a multi-layer convolutional sparse coding (ML-CSC) model has been proposed \cite{Papyan2017a}, which is proved to equal plain networks when propagates with the layered thresholding algorithm \cite{Papyan2017a}. This reveals that CNNs actually try to find the sparse coding or representation of input signals over a very special dictionary, which corresponds to the convolution operation. It computes the sparse vectors layer by layer, but does not recover all the vectors at once, which is computationally and conceptually challenging. The value of ML-CSC is not only giving us an understanding of CNNs, but also building a strict mathematical form which would provide a possibility of drawing into more mathematical tools to carry on strict theoretical analysis. Previous studies have theoretically proved that why ReLU behaves well in numerical experiments \cite{Nair2010}, \cite{Glorot2011}, what the feature maps computed in each layer represent, and what the meaning of the bias term (which is always added after convolution) is \cite{Papyan2017a}.

To solve the CSC model, layered basis pursuit (Layered BP) and a multi-layer version have been proposed \cite{Papyan2017a}, \cite{Sulam2019}. Layered BP considers the sparsity only in one layer, while the multi-layer version considers the sparsity with all layers together. The stability of Layered BP in noiseless and noisy regimes has been clarified in \cite{Papyan2017a}. A boundary has been proposed to measure the distance between the solution and the true underlying sparse coding \cite{Papyan2017a}. The convergence of the multi-layer version has been proved in \cite{Sulam2019}. The uniqueness of the sparsest representation and the conditions guaranteeing to find the true underlying one have been discussed in \cite{Papyan2017b}. Skipping a fixed number of spatial locations after convolution is a common step among practitioners of CNNs \cite{Krizhevsky2012}, \cite{Simonyan2014}. This can reduce the dimensions of the kernel maps throughout the layers, leading to computational benefits. More other theoretical benefits about this skill have been analyzed in \cite{Papyan2017a}.

At present, existing studies only established a preliminary connection between CSC and plain networks. The relationship between CSC and current popular architectures (e.g., ResNet, MSDNet) is still lacking. The roles of many key tricks (e.g., BN, Dropout) played in CSC are still not clear. We notice that neural networks with skip connections usually have better performance \cite{He2016}, \cite{Li2018}, \cite{Pelt2018}. Do the skip connections have any theoretical interpretation? Moreover, dilated convolution in MSDNet is also a powerful trick for extracting multi-scale features \cite{Pelt2018}. To better understand ResNet and MSDNet, we introduce a residual convolutional sparse coding model (Res-CSC) and a mixed-scale dense convolutional sparse coding model (MSD-CSC), which have close relationship with ResNet and MSDNet, respectively.

This paper is structured as follows. We first introduce notation and concepts in CNN and sparse coding in Section II. Second, we introduce the relationship between CNN and CSC in Section III. Third, we propose the layer-initializing question (LIQ) and the Res-CSC model in Section IV, and demonstrate ResNet is a special case of Res-CSC. Fourth, we propose the MSD-CSC model and establish the bridge between it and MSDNet in Section V. Through MSD-CSC, we give a theoretical understanding of superior performance of MSDNet in terms of dilated convolution and dense connection. Fifth, we implement two algorithms ISTA and FISTA to solve MSD-CSC in Section VI. Finally, we apply Res-CSC and MSD-CSC onto three typical datasets (CIFAR10, CIFAR100, SVHN) and compare it with several methods.

\section{notations and concepts}
\subsection{Convolution and Matrix Multiplication}
\begin{figure}[!t]
\centerline{\includegraphics[width=0.94\columnwidth]{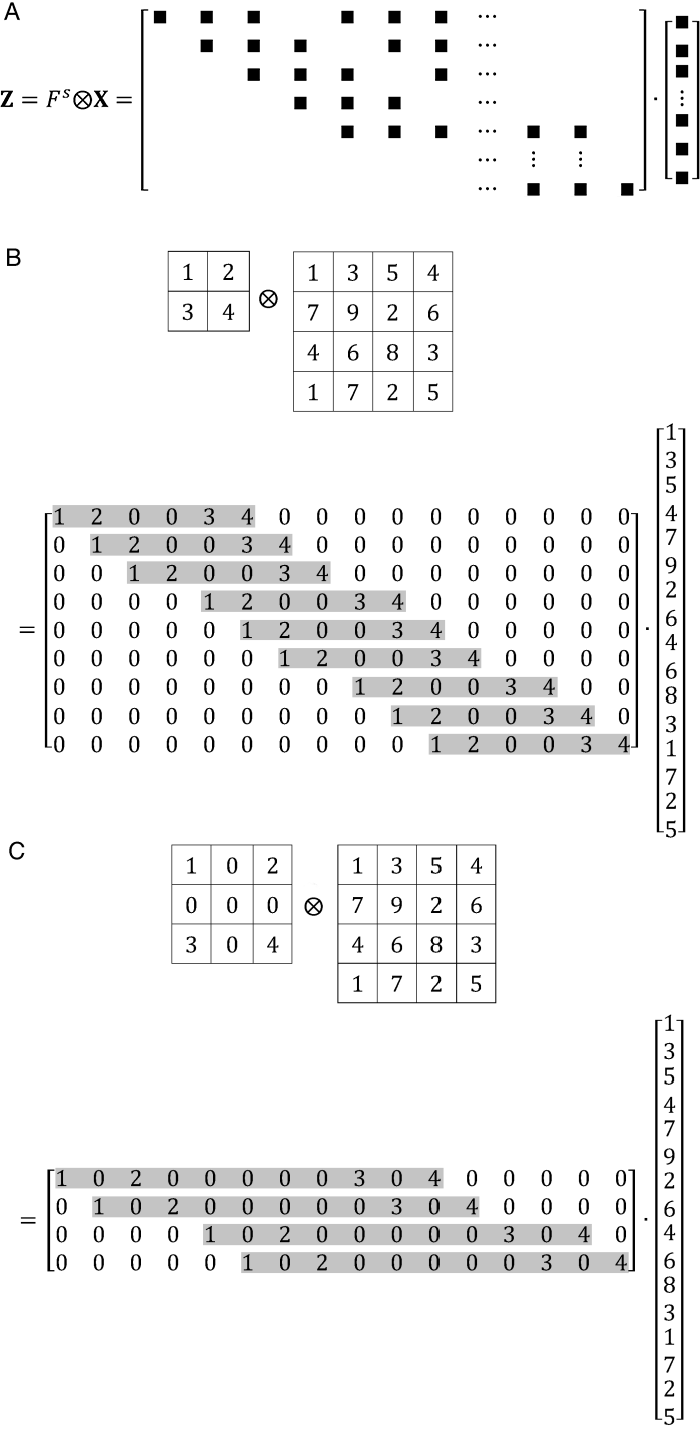}}
\caption{(A) The matrix-vector multiplication form of the convolution operation. (B) A $2\times2$ convolution kernel convolves a $4\times4$ image with dilation scale $s$=1. (C) A $2\times2$ convolution kernel convolves a $4\times4$ image with dilation scale $s$=2.}
\label{fig1}
\end{figure}

Convolution is a basic operation in CNN. It is used for extracting features in structured data, and has been widely used in computer vision. We denote an image with $m$ rows, $n$ columns and $c$ channels as $\textbf{X} \in R^{m \times n \times c}$. A dilated convolution kernel $F^{s}$ with dilation scales $s \in z^{+}$ \cite{Yu2015} convolves $\textbf{X}$ to produce a new feature map $\textbf{Z}$. This operation is denoted as $\textbf{Z}=F^{s} \otimes \textbf{X}$. Equally, this process can be written as a matrix multiplication (Fig. \ref{fig1}A). Simple examples with one channel and different dilation scales $s=1$ and $s=2$ are shown in Fig. \ref{fig1}B and Fig. \ref{fig1}C respectively. Here, we use the same symbol $F^{s}$ to represent the corresponding matrix, which has a special structure - a union of bars and circulant. $F^{s}$ is also called a convolutional matrix. Though convolution computes locally and scans globally, it is a linear transform about $\textbf{X}$. The structure of a convolutional matrix corresponds to the sparse connection and parameter sharing in convolution. These two characteristics decrease the number of parameters, which could overcome the overfitting issue to some extent.

\subsection{ResNet and MSDNet}
\begin{figure}[!t]
\centerline{\includegraphics[width=0.88\columnwidth]{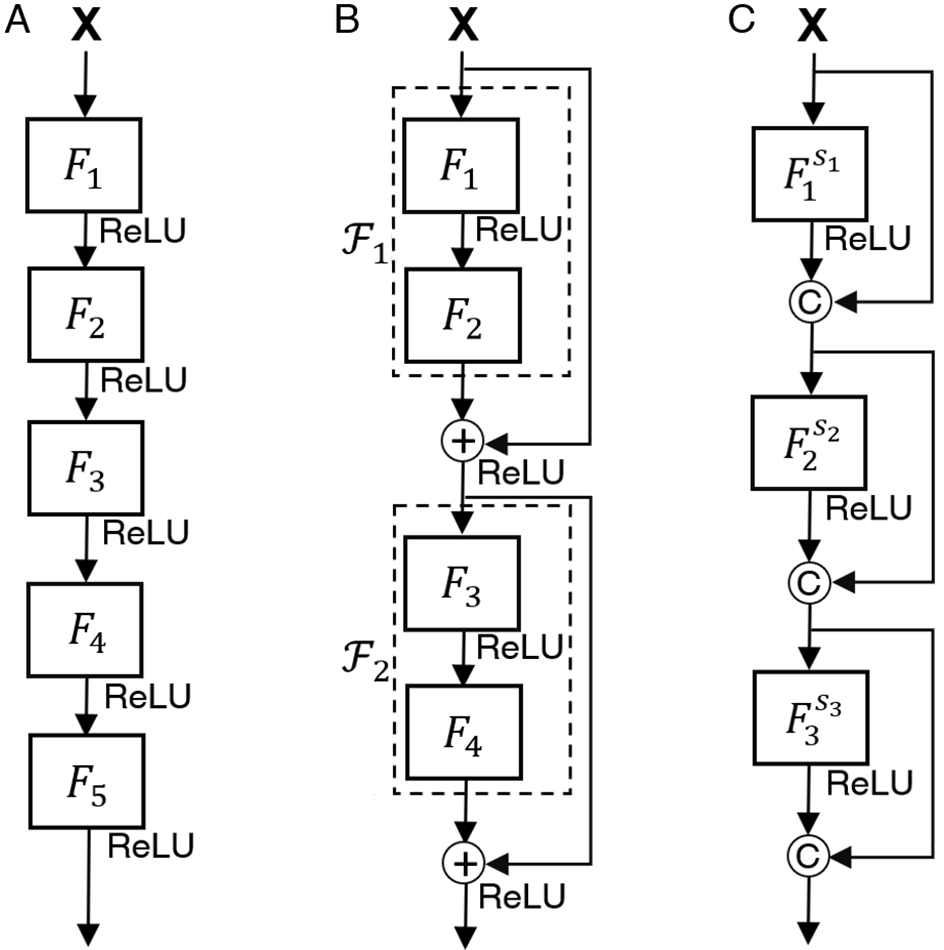}}
\caption{(A) A plain network of five layers by simply stacking each layer. (B) A ResNet of two layers. It assumes that the operation $\mathcal{F}$ does not change the shape of tensors. \textcircled{+} indicates the element-wise addition of the tensors of the current layer and previous ones. (C) A MSDNet of three layers. \textcircled{c} indicates the channel-wise concatenation of the tensors of the current layer and previous ones. }
\label{fig2}
\end{figure}

Compared with the traditional CNN (Fig. \ref{fig2}A), ResNet \cite{He2016} adds one operation - shortcut connection (Fig. \ref{fig2}B), and MSDNet \cite{Pelt2018} adds two operations - dilated convolution and dense connection (Fig. \ref{fig2}C). Shortcut connection directly adds feature maps after a transformation (Fig. \ref{fig2}B). This process can be written as below:
$$
\textbf{Z}_{i+1}=\mathcal{F}(\textbf{Z}_{i})+\textbf{Z}_{i}
$$
where $\mathcal{F}$ is a transformation and $\textbf{Z}_{i}$ is the $i$-th layer output.

Dilated convolutions with different dilation scales could acquire larger receptive field with fewer parameters, enabling feature extraction in a multi-scale manner. Dense connection gathers all feature maps before the current layer and computes new feature maps with them (Fig. \ref{fig2}C). This process can be written as below:
$$
\textbf{Z}_{i+1}=\sigma\left(F_{i+1}^{s_{i+1}}\left(\left\{\textbf{Z}_{0}, \textbf{Z}_{1}, \cdots, \textbf{Z}_{i}\right\}\right)+b_{i+1}\right)
$$
where $\sigma(\cdot)$ is the ReLU function, $\textbf{Z}_{0}$ is an input image, $\textbf{Z}_{i+1}$ is new feature maps, $\textbf{Z}_{i}$ is the $i$-th layer output, $\left\{\textbf{Z}_{0}, \textbf{Z}_{1}, \cdots, \textbf{Z}_{i}\right\}$ is feature maps which concatenates $\textbf{Z}_{0}$, $\textbf{Z}_{1}$, $\cdots$, $\textbf{Z}_{i}$, and $b_{i+1}$ is a bias term.

In this paper, a MSDNet with $k$ layers is represented as:
$$
MSD^{k}=\left\{\left(F_{1}^{s_{1}}, b_{1}\right),\left(F_{2}^{s_{2}}, b_{2}\right), \cdots,\left(F_{k}^{s_{k}}, b_{k}\right)\right\}
$$
where $\left(F_{i}^{s_{i}}, b_{i}\right)$ denotes the convolution kernels and the bias $b_{i}$ in the $i$-th layer. Note that $b_{i}$ is a vector recording biases corresponding to convolution kernels in $F_{i}^{s_{i}}$.
Here, we use $w$ to denote the number of convolution kernels in each layer and $d$ to denote the number of layers.

\subsection{Sparse Coding}
Let $X$ denote a signal vector. In sparse coding, one assumes that it can be represented as a linear combination of a few columns of a dictionary matrix $D$:
$$X=D \cdot \Gamma $$
where $\Gamma$ is the coefficient vector of the linear combination, and it is called a coding under the dictionary $D$.

In general, $\Gamma$ is not unique because the number of columns in $D$ is usually greater than the number of rows in $D$. One expects to find a special coding with a small number of non-zeros in all of the solutions. In addition, the signal $X$ always has noise, we don't need to reconstruct $X$ exactly. Finally, sparse coding can be formulated into the following optimization problem \cite{Donoho2003}, \cite{Tropp2004}, \cite{Rubinstein2009}:
$$(\mathrm{P} 0): \min _{\Gamma} \frac{1}{2}\|X-D \Gamma\|_{2}^{2}+\beta\|\Gamma\|_{0}$$
where $\beta$ is a regularization parameter to balance the reconstruction error of the signal $X$ and the sparsity of $\Gamma$. $\|\Gamma\|_{0}$ denotes the number of non-zero entries in $\Gamma$.

The problem (P0) is NP-hard because of the second term $\|\Gamma\|_{0}$ \cite{Natarajan1995}. Fortunately, it has been proved that the problem (P0) can be relaxed into the following format \cite{Candes2004}:
$$
(\mathrm{P} 1): \min _{\mathrm{\Gamma}} \frac{1}{2}\|X-D \Gamma\|_{2}^{2}+\beta\|\Gamma\|_{1}
$$
Here the problem (P1) is called the Lasso \cite{Tibshirani1996} or BP problem \cite{Donoho2003}, \cite{Chen1998}, \cite{Chen2001}, \cite{Tropp2006} in different fields. Moreover, the problem (P1) can be solved using the popular iterative soft thresholding algorithm (ISTA).
Its update formula can be formulated as follows:
\begin{equation}
\Gamma^{k+1}=S_{\frac{\beta}{L}}\left(\Gamma^{k}-\frac{1}{L}\left(-D^{T} X+D^{T} D \Gamma^{k}\right)\right)
\end{equation}
where $\Gamma^{k}$ denotes the coding in the $k$-th iteration. The smallest Lipschitz constant ($L$) of the gradient of $f(\Gamma)=\frac{1}{2}\|X-D \Gamma\|_{2}^{2}$ is $2\lambda_{\max }\left(D^{T} D\right)$, where $\lambda_{\max }\left(D^{T} D\right)$ denotes the maximum eigenvalue of $D^{T} D$.
The soft thresholding operator $S_{b}(\cdot)$ is defined as below:
$$
S_{b}(z)=\left\{\begin{array}{ll}{z+b,} & {z<-b} \\ {0,} & {-b \leq z \leq b} \\ {z-b,} & {z > b}\end{array}\right.
$$

Is the sparsest representation for the problem (P1) unique? \textbf{Lemma 1} gives the answer. \\
{\bf Lemma 1 \cite{Donoho2003}, \cite{Tropp2004}, \cite{Donoho2005}:} The sparsest  representation is unique if the number of non-zeros in the underlying sparsest representation for the problem (P1) is not too high and in particular less than $\frac{1}{2}\left(1+\frac{1}{\mu(D)}\right)$. Here, $\mu(D)$ is defined as the maximal inner product of two columns extracted from $D$. This can be formally written as below:$$
\mu(D)=\max _{i \neq j}\left|d_{i}^{T} d_{j}\right|
$$
where $d_i$ is assumed to be normalized to unit length.

\section{The relationship between CSC and CNN}
\subsection{Nonnegative Sparse Coding}
Let's consider a signal $X=D \Gamma$. Naturally, $\Gamma$ can be split into its positive one $\Gamma_{P}$ and negative one $\Gamma_{N}$. Then $X$ can be equally written as:
$$
X=D \Gamma_{P}+(-D)\left(-\Gamma_{N}\right)
$$
Obviously, if we change the dictionary into $[D,-D]$, then the corresponding sparse coding is $\left[\Gamma_{P},-\Gamma_{N}\right]^{T}$. Notice both $\Gamma_{P}$ and $- \Gamma_{N}$ are nonegative. Therefore, every sparse coding can always be converted into nonnegative sparse coding.

\subsection{Soft Non-negative Thresholding Operator}
For a nonnegative sparse coding problem, one only needs to consider a nonnegative sparse coding. So, one can define the soft non-negative thresholding operator $S_{b}^{+}(\cdot)$ based on the soft thresholding operator:
$$
S_{b}^{+}(z)=\left\{\begin{array}{ll}{0,} & {z \leq b} \\ {z-b,} & {z > b}\end{array}\right.
$$
It is obvious that the soft non-negative thresholding operator is equivalent to the ReLU:\begin{equation}
S_{b}^{+}(z)=\max (z-b, 0)=\operatorname{ReLU}(z-b)
\end{equation}
where $b$ is a bias term. According to Eq. (1), $b$ depends on $\beta$ and the Lipschitz constant $L$ in the problem (P1). In other words, $\beta$ is a hyper-parameter in sparse coding, but it becomes a parameter in neural networks via Eq. (2).

\subsection{ML-CSC}
The ML-CSC model is formulated as follows \cite{Papyan2017a}:
\begin{center}
$X=D_{1} \Gamma_{1}$\\
$\Gamma_{1}=D_{2} \Gamma_{2}$\\
$\cdots\cdots$\\
$\Gamma_{k-1}=D_{k} \Gamma_{k}$
\end{center}
where $X$ is the input signal (e.g., an image), and $\left\{D_{i}\right\}$ is a set of special dictionaries. Each $D_{i}$ is a transpose of a convolutional matrix. ML-CSC encodes signals layer by layer:
\begin{center} $\Gamma_{1}$ reconstructs $X$ via dictionary $D_{1}$\\$\Gamma_{2}$ reconstructs $\Gamma_{1}$ via dictionary $D_{2}$\\......\\$\Gamma_{k}$ reconstructs $\Gamma_{k-1}$ via dictionary $D_{k}$ \end{center}
The $i$-th layer in ML-CSC can be described as a Lasso problem:$$
(\mathrm{P}2): \min _{\Gamma_{i}} \frac{1}{2}\left\|\Gamma_{i-1}-D_{i} \Gamma_{i}\right\|_{2}^{2}+\beta\left\|\Gamma_{i}\right\|_{1}
$$One can use Eq. (1) to compute the sparse coding in every layer. When $\left\{D_{i}\right\}$ is known, one sets $\Gamma^{0}=0$, and Eq. (1) becomes:
\begin{equation}
\Gamma^{1}=S_{\frac{\beta}{L}}\left(\frac{1}{L}\left(D^{T} X\right)\right)
\end{equation}
One can update $\Gamma$ only once with Eq. (3) and then obtain the layered thresholding algorithm with $k$ layers (Algorithm 1).

\renewcommand{\algorithmicrequire}{ \textbf{Input:}} 
\renewcommand{\algorithmicensure}{ \textbf{Output:}} 

\begin{algorithm}
\caption{The layered thresholding algorithm}

\begin{algorithmic}
\REQUIRE ~~\\ 
Signal $X$\\
A set of special dictionaries $\left\{D_{i}\right\}$\\
A set of special thresholding $\left\{b_{i}\right\}$\\
Thresholding operator $P \in\left\{S, S^{+}\right\}$
\ENSURE ~~\\ 
Encoding signals $\left\{\widehat{\Gamma}_{i}\right\}$, $i=1,2,\cdots,k$
\STATE 1. $\hat{\Gamma}_{0} \leftarrow X$
\STATE 2. for $i=1:k$ do:
\STATE 3. \quad$\widehat{\Gamma}_{i} \leftarrow P_{b_{i}}\left(D_{i}^{T} \widehat{\Gamma}_{i-1}\right)$
\end{algorithmic}
\end{algorithm}

\begin{figure*}[!t]
\centerline{\includegraphics[width=1.6\columnwidth]{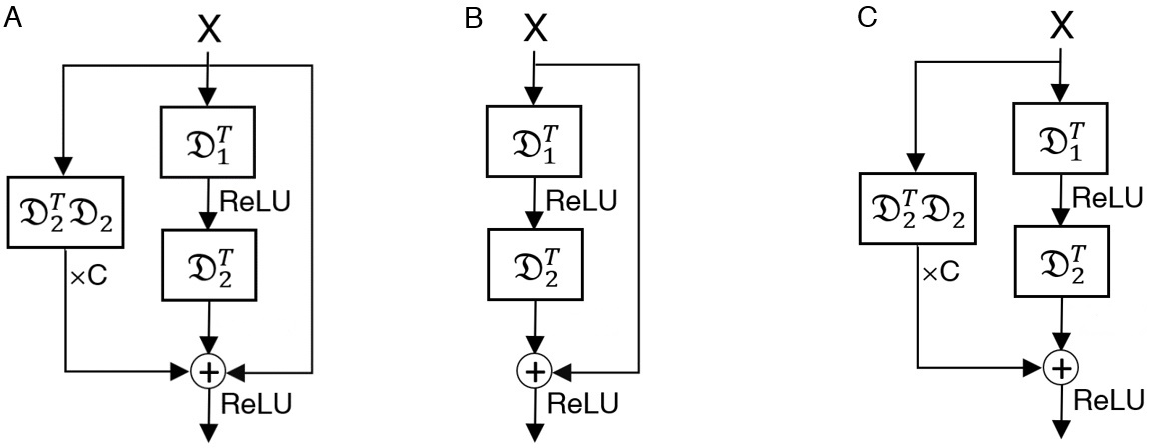}}
\caption{(A) A component of Res-CSC with two continuous layers. (B) A component of ResNet with two continuous layers. It can be seen as a simplification of (A). (C) Another way to simplify (A) (denoted as Res-CSC-Simplified). Here the $\mathfrak{D_1}$ and $\mathfrak{D_2}$ are the dictionaries of the two continuous layers respectively.}
\label{fig3}
\end{figure*}

According to the relationship between convolution and matrix multiplication, it is obvious that using this layered thresholding algorithm to solve ML-CSC is equivalent to the forward pass of the plain networks (Fig. \ref{fig2}A) \cite{Papyan2017a}. So, the final sparse coding $\Gamma_{k}$ in ML-CSC corresponds to the final feature map in CNN. Here one sets $\Gamma^{0}=0$ and only updates $\Gamma$ once in each layer. This strategy not only achieves the equivalence between ML-CSC and the plain networks, but also improve the computational efficiency since two terms, $\Gamma^{k}$ and $D^{T}D\Gamma^{k}$, can be ignored.

A natural question is how close between the true solution $\Gamma_{i}$ and the estimate $\widehat{\Gamma}_{i}$ solved by Algorithm 1. \textbf{Lemma 2} answers this question. \\
{\bf Lemma 2 \cite{Papyan2017a}}: Suppose a signal $X$ with noise $E$, where $\|E\|_{2} \leq \varepsilon_{0}$. For all $1 \leq i \leq k$, if$$
\left\|\Gamma_{i}\right\|_{0, \infty}^{s} \leq \frac{1}{2}\left(1+\frac{1}{\mu\left(D_{i}\right)}\right)
$$then$$
\left\|\Gamma_{i}-\hat{\Gamma}_{i}\right\|_{2}^{2} \leq \varepsilon_{0}^{2} \prod_{k=1}^{i} \frac{4}{1-\left(2\left\|\Gamma_{k}\right\|_{0, \infty}^{s}-1\right) \mu\left(D_{k}\right)}
$$
where $\left\|\Gamma_{i}\right\|_{0, \infty}^{s}$ is defined to be the maximal number of non-zeros in stripe of length $(2n-1)m$ extracted from $\Gamma_{i}$ ($m$ denotes the number of convolution kernels in the $i$-th layer and $n$ denotes the length of bars in the $i$-th layer) (Fig. \ref{fig1}).

Until now, ML-CSC has been connected to the plain network. Intriguingly, we think three factors in each layer of it including the initialization, the dictionary design and the number of iterations greatly affect its performance.
Inspired by these considerations, in the next three sections IV, V and VI, we propose Res-CSC and MSD-CSC, and a forward propagation algorithm with unfolding (iterate more than once), respectively.

\section{LIQ AND RES-CSC}
According to the above analysis, the forward pass of plain networks can be explained as solving Eq. (1) with initialization $\Gamma^{0}=0$ in each layer. This dramatically improves the computational efficiency. However, this naive setting causes serious training limits in much deeper networks due that its iteration time in each layer is set to be one, which causes the accumulation of errors. Thus, we propose a fundamental question about this initialization:

{\bf Layer-Initializing Question (LIQ)}: In ML-CSC, Eq. (1) iterates once in each layer. Under this condition, can we design a proper initialization for $\Gamma_i^{0}$ of the $i$-th layer 
to approach the optimal sparse coding $\Gamma_{i}$?

To give a solution to LIQ, we modify the ML-CSC model. In principle, ML-CSC computes the sparse coding layer by layer, and such a representation becomes sparse gradually. Thus, an intuitive idea is to set $\Gamma^{0}$ equals the input of a former layer. In the following, we use the input of the closest layer to the current one as the initialization (denoted as $\Gamma^{0}=X_{-1}$). We change the forward propagation rule of each layer partially in ML-CSC to reduce the accumulation error, and keep part of the rule to enhance the computational efficiency. 
Specifically, every two layers use this new initialized setting once. The first one adopts Eq. (3) to obtain the output, and the second one employs the following update rule based on Eq. (1):
\begin{equation}
\Gamma^{1}=S_{\frac{\beta}{L}}\left(\frac{1}{L} D^{T} X+X_{-1}-\frac{1}{L} D^{T} D X_{-1}\right)
\end{equation}
Let $\mathfrak{D}=\frac{1}{L} D$, $c=-L$. Eq. (4) becomes:
\begin{equation}
\Gamma^{1}=S_{\frac{\beta}{L}}\left(\mathfrak{D}^{T} X+X_{-1}+\mathbf{c} \cdot \mathfrak{D}^{T} \mathfrak{D} X_{-1}\right)
\end{equation}
Now, we obtain a new optimization rule following the same mode of ML-CSC. In contrast, its forward propagation implements Eq. (3) and Eq. (5) alternately (Fig. \ref{fig3}A). Notice that Eq. (5) becomes the forward propagation of ResNet exactly when we ignore the term $\mathrm{c} \cdot \mathfrak{D}^{T} \mathfrak{D} X_{-1}$. For convenience, we name the ML-CSC updated with this new rule as Res-CSC. The classical ResNet can be seen as a special case of Res-CSC (Fig. \ref{fig3}B), which gives an approximate solution to the LIQ. But it has not been formally proposed before that LIQ is a key of the optimization problem behind degradation phenomenon \cite{He2016}. Similarly, we can obtain another approximate update rule by ignoring the term $X_{-1}$ (Fig. \ref{fig3}C). 

The relationship between Res-CSC and ResNet draws our attention to the error-tolerant, which is an important characteristic of CNN \cite{Gupta2015}, \cite{Courbariaux2016}. Being the approximation of Res-CSC, ResNet has presented this characteristic in many applications. For a more general propagation rule, Res-CSC can overcome the training difficulty when networks come to hundreds of layers. 
In the experiment section, numerical tests on Res-CSC, ResNet and Res-CSC-Simplified using the three typical datasets demonstrate that the terms $X_{-1}$ and $\mathrm{c} \cdot \mathfrak{D}^{T} \mathfrak{D} X_{-1}$ in Eq. (5) indeed play a key role. This reveals that LIQ is the underlying mathematical key behind the degradation phenomenon in plain networks. Coincidentally, ResNet addresses it by introducing such non-zero initializations.

\section{MSD-CSC AND THEORETICAL ANALYSIS}
\subsection{MSD-CSC}
Inspired by the layered thresholding algorithm designed for ML-CSC, we attempt to describe the dilated convolution and dense connection of MSDNet in a sparse coding view via modifying the structure of dictionaries. In MSDNet, each layer uses all the previous feature maps to compute its layer output. 
This leads to the following CSC model:
\begin{center}
$X=D_{1}^{s_{1}} \Gamma_{1}$\\$\Gamma_{1}=D_{2}^{s_{2}} \Gamma_{2}$\\......\\$\Gamma_{k-1}=D_{k}^{s_{k}} \Gamma_{k}$
\end{center}
where $X$ is the input signal (e.g., an image), $D_{i}^{s_{i}}=\left[\begin{array}{ll}{\mathrm{I}} & {\left(F_{i}^{s_{i}}\right)^{T}}\end{array}\right]$, and $I$ is an identity matrix. The Lasso problem in the $i$-th layer is formulated as:$$
\min _{\Gamma_{i}} \frac{1}{2}\left\|\Gamma_{i-1}-D_{i}^{s_{i}} \Gamma_{i}\right\|_{2}^{2}+\beta\left\|\Gamma_{i}\right\|_{1}
$$
We call this the MSD-CSC model, and denote it as:$$
MSDCSC^{k}=\left\{\left(D_{1}^{s_{1}}, \beta_{1}\right),\left(D_{2}^{s_{2}}, \beta_{2}\right), \cdots,\left(D_{k}^{s_{k}}, \beta_{k}\right)\right\}
$$
where $D_{i}^{s_{i}}$ is the dictionary and $\beta_{i}$ is the regularization parameter in the $i$-th layer. Is MSD-CSC equivalent to MSDNet?

\subsection{Theoretical Analysis}
{\bf Proposition 1}: For a given MSDNet, there exists a MSD-CSC model, which is equivalent to MSDNet when propagates with the layered thresholding algorithm.  \\
{\bf Proof}: For a given MSDNet:
$$
MSD^{k}=\left\{\left(F_{1}^{s_{1}}, b_{1}\right),\left(F_{2}^{s_{2}}, b_{2}\right), \cdots,\left(F_{k}^{s_{k}}, b_{k}\right)\right\}
$$
Let's define a corresponding MSD-CSC model:
$$
MSDCSC^{k}=\left\{\left(D_{1}^{s_{1}}, \beta_{1}\right),\left(D_{2}^{s_{2}}, \beta_{2}\right), \cdots,\left(D_{k}^{s_{k}}, \beta_{k}\right)\right\}
$$
where $\beta_{i}=\left(0,\cdots, 0, L_{i}b_{i}\right)^{T}$. $L_{i}$ is the Lipschitz constant in the $i$-th layer. According to the layered thresholding algorithm:
$$
\begin{aligned}
\widehat{\Gamma}_{i} &=S^+_{\overline{b}_{i}}\left(\left(D_{i}^{s_{i}}\right)^{T} \hat{\Gamma}_{i-1}\right) \\ &=S^+_{\overline{b}_{i}}\left(\left[\begin{array}{c}{I} \\
{F_{i}^{s_{i}}}\end{array}\right] \hat{\Gamma}_{i-1}\right) \\
&=S^+_{\overline{b}_{i}}\left(\left[\begin{array}{c}{\hat{\Gamma}_{i-1}} \\
{F_{i}^{s_{i}}\widehat{\Gamma}_{i-1}} \end{array}\right]\right) \\
&=\operatorname{ReLU}\left(\left[\begin{array}{c}{\hat{\Gamma}_{i-1}} \\ {F_{i}^{s_{i}}\widehat{\Gamma}_{i-1}+b_{i}}\end{array}\right]\right) \\
&=\left[\begin{array}{c}{\hat{\Gamma}_{i-1}} \\ {Z_{i}}\end{array}\right] \\
&=\text{concatenate}\left(\hat{\Gamma}_{i-1}, Z_{i}\right)
\end{aligned}
$$
where $i=1,\cdots, k$, $\overline{b}_{i}=\frac{\beta_{i}}{L_{i}}=(0,\cdots,0,b_{i})$. We can observe that the features before the $i$-th layer are kept in the $i$-th feature as the input of the next layer. This indicates that the propagating rule of $MSDCSC^{k}$ is equivalent to that of $MSD^{k}$. $\blacksquare$

From {\bf Proposition 1}, we can see that MSDNet is just a special case of a corresponding MSD-CSC model. Next, we will proof that the coding performance of MSD-CSC is better than that of ML-CSC. This attributes to two operations: dilation convolution and dense connection. In context of CSC, the dilation convolution affects the structure of the convolutional matrices. From Fig. \ref{fig1}B and Fig. \ref{fig1}C, we can see that $\mu(D)$ is relatively smaller in the convolutional matrix corresponding to the dilated convolutional kernel compared with that without dilation ($\mu=0.47$ in Fig. \ref{fig1}B and $\mu=0$ in Fig. \ref{fig1}C). According to {\bf Lemma 1}, a dictionary (notice that the identity matrices in MSD-CSC don't affect $\mu$) with smaller $\mu$ tends to ensure that the sparsest representation is unique. According to {\bf Lemma 2}, a dictionary with smaller $\mu$ will make the estimation $\widehat{\Gamma}_{i}$ closer to the true solution $\Gamma_{i}$ when solved with Algorithm 1.

Next we will explore how MSD-CSC benefits from the dense connection. In sparse coding, larger $\beta$ leads to sparser representation, but sparser representation may lead to the loss of information according to the Lasso form. Sparsity and loss of information are contradictory. Sometimes, an unsuitable $\beta$ can led to a very unreasonable solution. For example, let's $D=\left[\begin{array}{lll}{0.1} & {0.1} & {0.2} \\ {0.1} & {0.2} & {0.1}\end{array}\right]$ and $\beta=\frac{1}{3}$. According to the update rule Eq. (1), the coding of the signal $X=(1,1)^{T}$ over the $D$ is $\Gamma=(0,0,0)^{T}$. It is obviously unreasonable that all the information are lost. This ill-conditioned case leads to the following definition. \\
\textbf{Definition 1:} For the encoding process in the $i$-th layer: $\Gamma_{i-1}=D_{i}^{s_{i}} \Gamma_{i}$, the corresponding Lasso problem is:
$$
\min _{\Gamma_{i}} \frac{1}{2}\left\|\Gamma_{i-1}-D_{i}^{s_{i}} \Gamma_{i}\right\|_{2}^{2}+\beta\left\|\Gamma_{i}\right\|_{1}
$$
Let $\xi=\left(\xi_{1}, \xi_{2}, \cdots, \xi_{n}\right)=D_{i}^{s_{i}} \Gamma_{i}$, which is used to reconstruct the signal $\Gamma_{i-1}$. If the $j$-th dimension satisfy:
$$\left|\xi_{j}-\Gamma_{i-1}^{(j)}\right|>2 \beta$$
where $\Gamma_{i-1}^{(j)}$ denotes the value in the $j$-th dimension of $\Gamma_{i-1}$. The reconstruction is considered as unsuccess in the $j$-th dimension, otherwise success.

\begin{figure}[!t]
\centerline{\includegraphics[width=0.88\columnwidth]{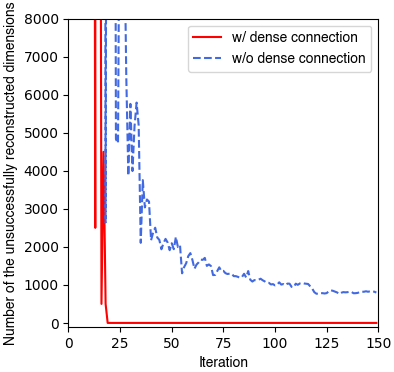}}
\caption{The number of unsuccessfully reconstructed dimensions in the original signal after each iteration identified by the models with (MSD-CSC) or without (ML-CSC) the dense connection.}
\label{fig4}
\end{figure}

Let's conduct a simulated experiment on ML-CSC first. The simulation data is of length 100 and generated by adding the Gaussian noise to 100 different data centers, which represent 100 different classes. We generate 10000 training data (each class has 100 data points) and 2000 testing data. We use a ML-CSC with two hidden layers. After each iteration, we compute the regularization parameter $\beta$, reconstruct the input signal $X$ and count the number of dimensions which fails to be reconstructed (i.e., unsuccess) (Fig. \ref{fig4}). We can see that the number of unsuccessfully reconstructed dimensions decreases at the beginning of the training process and becomes stable after some iterations. Finally, there still exists some dimensions which can not be reconstructed successfully.

The theorem below could show us that MSD-CSC is better than ML-CSC under this situation. \\
{\bf Theorem 1}: For the Lasso problem in each layer, the performance of MSD-CSC is better than that of ML-CSC. \\
{\bf Proof}: In ML-CSC, the coding in the $i$-th layer is $\Gamma_{i-1}=D_{i} \Gamma_{i}$, and the corresponding Lasso problem is: $$
\min _{\Gamma_{i}} \frac{1}{2}\left\|\Gamma_{i-1}-D_{i} \Gamma_{i}\right\|_{2}^{2}+\beta\left\|\Gamma_{i}\right\|_{1}
$$
Let $\xi=\left(\xi_{1}, \xi_{2}, \cdots, \xi_{n}\right)=D_{i} \Gamma_{i}$, $$
f_{ML-CSC}=\frac{1}{2}\left\|\Gamma_{i-1}-D_{i} \Gamma_{i}\right\|_{2}^{2}+\beta\left\|\Gamma_{i}\right\|_{1}
$$
In MSD-CSC, the coding in the $i$-th layer is
$\Gamma_{i-1}=D_{i}^{s_{i}} \Gamma_{i}=\left[\begin{array}{ll}{\mathrm{I}} & {\left.\left(F_{i}^{s_{i}}\right)^{T}\right] \Gamma_{i}}\end{array}\right.,$ and the corresponding Lasso problem is
$$
\min _{\Gamma_{i}} \frac{1}{2}\left\|\Gamma_{i-1}-D_{i}^{s_{i}} \Gamma_{i}\right\|_{2}^{2}+\beta\left\|\Gamma_{i}\right\|_{1}
$$
Let $f_{MSD-CSC}=\frac{1}{2}\left\|\Gamma_{i-1}-D_{i}^{s_{i}} \eta\right\|_{2}^{2}+\beta\|\eta\|_{1}$.

(i) In ML-CSC, if the reconstructions in all the dimensions are considered as success, let $\eta=\left(0,\cdots, 0 | \Gamma_{i}\right)^{T}$. Then we can obtain:$$
\begin{aligned} f_{MSD-CSC} &=\frac{1}{2}\left\|\Gamma_{i-1}-D_{i}^{s_{i}} \eta\right\|_{2}^{2}+\beta\|\eta\|_{1} \\ &=\frac{1}{2}\left\|\Gamma_{i-1}-\left(F_{i}^{s_{i}}\right)^{T} \Gamma_{i}\right\|_{2}^{2}+\beta\left\|\Gamma_{i}\right\|_{1} \\ &=f_{M L-C S C} \end{aligned}
$$
(ii) 	In ML-CSC, if the reconstruction in the $j$-th dimension is considered as unsuccess, let $$
\eta=\left(0, \cdots, \Gamma_{i-1}^{(j)}-\xi_{j}, 0, \cdots, 0 | \Gamma_{i}\right)^{T}
$$where $\Gamma_{i}$ is the solution of ML-CSC (Fig. 5). Then, we can obtain:$$
\begin{aligned} &\quad\frac{1}{2}\left\|\Gamma_{i-1}-D_{i}^{s_{i}} \eta\right\|_{2}^{2}+\beta\|\eta\|_{1} \\
                     &=\frac{1}{2}\left\|\Gamma_{i-1}-\left[\begin{array}{cc}{I} & {\left(F_{i}^{s_{i}}\right)^{T}}\end{array}\right] \eta\right\|_{2}^{2}+\beta\|\eta\|_{1} \\
                     &=\frac{1}{2}\left\|\Gamma_{i-1}-\left(F_{i}^{s_{i}}\right)^{T} \Gamma_{i}\right\|_{2}^{2}-\frac{1}{2}\left(\xi_{j}-\Gamma_{i-1}^{(j)}\right)^{2}+\\
                     &\quad\quad\quad\quad\quad\quad\quad\quad\quad\quad\quad\quad\quad\quad\quad\quad \beta\left\|\Gamma_{i}\right\|_{1}+\beta\left|\xi_{j}-\Gamma_{i-1}^{(j)}\right|\\
                    &=f_{ML-CSC}-\frac{1}{2}\left(\xi_{j}-\Gamma_{i-1}^{(j)}\right)^{2}+\beta\left|\xi_{j}-\Gamma_{i-1}^{(j)}\right|<f_{ML-CSC}
                    \end{aligned}
$$
This indicates that MSD-CSC has a better solution for the Lasso problems in each layer than ML-CSC. $\blacksquare$

\begin{figure}[!t]
\centerline{\includegraphics[width=0.95\columnwidth]{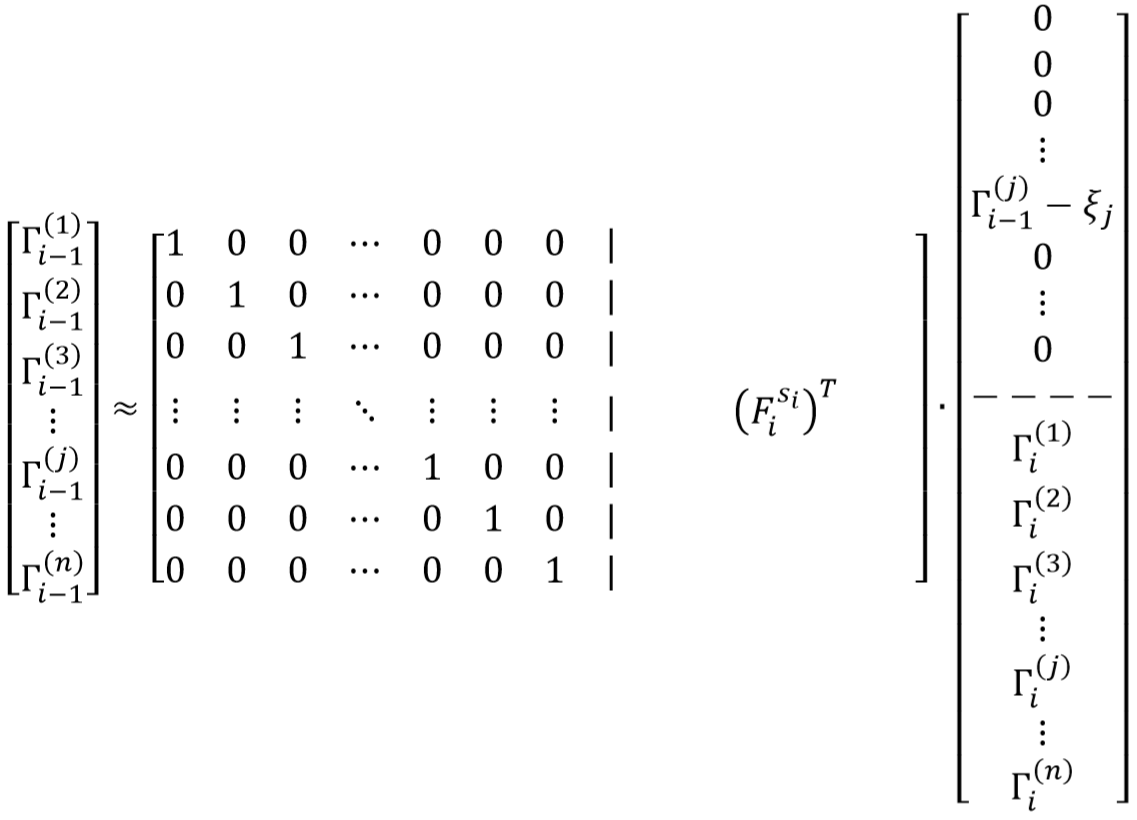}}
\caption{Illustration of a sparse coding scheme and the recovery role of the identity matrix of the dictionary in the $i$-th layer of MSD-CSC. Assume the signal in the ($i$-1)-th layer is $n$.}
\label{fig5}
\end{figure}

\begin{figure*}[!t]
\centerline{\includegraphics[width=1.70\columnwidth]{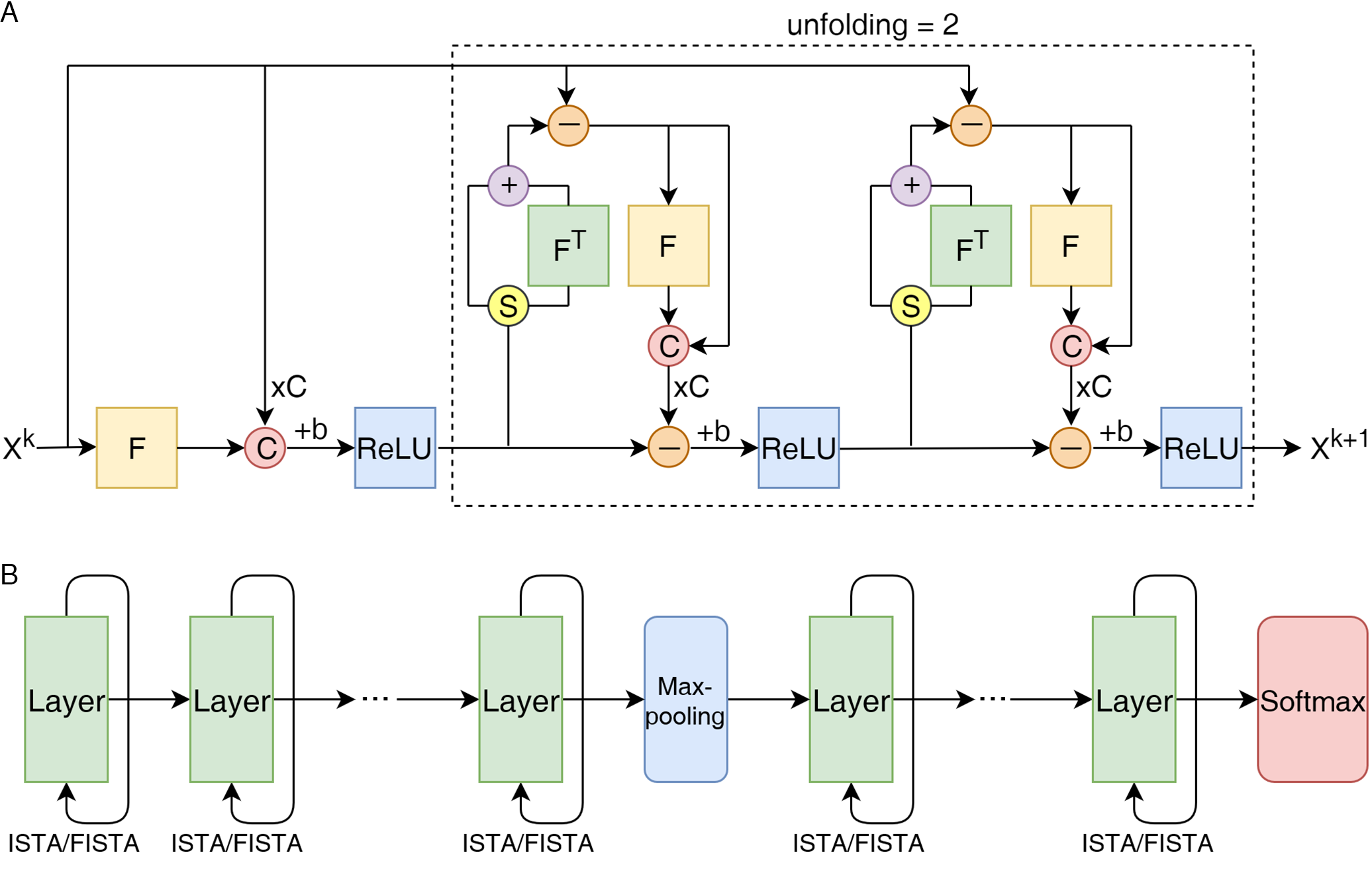}}
\caption{(A) An illustration shows MSD-CSC ISTA with unfolding = 2. $X_{k}$ denotes the signal input the $k$-th layer. $F$ denotes convolution and $F^{T}$ denotes multiply with the transpose of convolutional matrix, \textcircled{c} denotes concatenate operation, \textcircled{s} denotes split channels, here we split last $w$ channels, $w$ is the number of convolution kernels in this layer. $+$, $-$, $×$ and ReLU represents their literal meaning. (B) An architecture designed for a classification task. Tensor flow in each layer is illustrated in (A) with ISTA as the forward propagation algorithm.}
\label{fig6}
\end{figure*}

Let's further conduct a simulated experiment on MSD-CSC. Note that the bias term of MSD-CSC has been updated to ensure that it has the same regularization parameter as ML-CSC. Based on the simulated experiment, we can see that the number of unsuccessfully reconstructed dimensions decreases rapidly compared with that of ML-CSC and finally stabilizes at zero (Fig. \ref{fig4}). This means that the unsuccessfully reconstructed dimensions in ML-CSC can been recovered by MSD-CSC due to the identity matrix (corresponding to the dense connection in MSDNet).

In {\bf Theorem 1}, we only consider the situation that one specific dimension is considered as unsuccess. Actually, the proof can be extended to all the dimensions that are considered as unsuccess. In this case, $f_{MSD-CSC}$ could be much smaller than $f_{ML-CSC}$. Besides, in this proof, we still assume the balance coefficient $\beta$s in ML-CSC and the corresponding MSD-CSC are equivalent. This assumption is reasonable under the view of CSC since MSD-CSC and the corresponding ML-CSC can be seen as two arrays of Lasso problems with the same coefficient $\beta$. But from the view of neural networks, there doesn't exist the concept of regularization parameter $\beta$. Thus, the corresponding ML-CSC should be considered as having the same bias as MSD-CSC. How should we understand the better performance under this new assumption?  \\
{\bf Lemma 3:} For a matrix $A \neq 0$, let's assume the matrix $A A^{T}$ has eigenvalues $\lambda_{1}, \cdots, \lambda_{n}$. As a result, the matrix
$$
B=
\left(\begin{array}{l}{I} \\
{A}\end{array}\right) \cdot\left(\begin{array}{ll}{I} & {A^{T}}\end{array}\right)=\left(\begin{array}{ll}{I}&{A^{T}} \\ {A} & {A A^{T}}
\end{array}\right)
$$
has eigenvalues $0,\cdots, 0, \lambda_{1}+1, \cdots, \lambda_{n}+1$. Here the number of zeros is equivalent to the number of columns of $A$.  \\
{\bf Proof}: Assume $X$ is an eigenvector of $A A^{T}$ corresponding to the eigenvalue $\lambda$. Let's consider the vector $X^{\prime}=\left(\frac{1}{\lambda} A^{T} X, X\right)^{T} \neq 0$ and the following equation
$$
\left(\begin{array}{cc}{I} & {A^{T}} \\ {A} & {A A^{T}}\end{array}\right) \cdot \left(\begin{array}{c}{\frac{1}{\lambda} A^{T} X} \\ {X}\end{array}\right)=\left(\begin{array}{c}{\frac{1}{\lambda} A^{T} X+A^{T} X} \\
{X+\lambda X}\end{array}\right)
$$
$$\quad\quad\quad\quad\quad\quad\quad\quad\quad\quad\quad\quad\ =(\lambda+1)\left(\begin{array}{c}{\frac{1}{\lambda} A^{T} X} \\
{X}\end{array}\right)
$$
Obviously, $B$ has an eigenvalue $\lambda_{i}+1$ and $X^{\prime}$ is a corresponding eigenvector. So, $B$ has eigenvalues $\lambda_{1}+1,\cdots,\lambda_{n}+1$. Let's further consider the trace of $B$, $\operatorname{tr}(B)=\operatorname{tr}(I)+\operatorname{tr}\left(A A^{T}\right)=\lambda_{1}+\lambda_{2}+\cdots+\lambda_{n}+n$. Then, the sum of the rest eigenvalues is zero because the trace of a matrix is equivalent to the sum of all eigenvalues. In addition, all the eigenvalues of $B$ are nonnegative (A nonzero matrix with the form of $M M^{T}$ has nonnegative eigenvalues). Taken together, all the rest eigenvalues of $B$ are zeros. $\blacksquare$

In MSD-CSC, the Lipschitz constant
$$
L_{MSD-CSC}=2 \lambda_{\max }\left(\left(\begin{array}{c}{I} \\ {F_{i}^{s_{i}}}\end{array}\right)\left(\begin{array}{cc}{I} & {F_{i}^{s_{i} T}}\end{array}\right)\right)
$$
According to {\bf Lemma 3}, $L_{MSD-CSC}=L_{ML-CSC}+2$, where $L_{ML-CSC}$ is the Lipschitz constant of the corresponding ML-CSC. According to Eq. (1) and Eq. (2), $\beta=bL$. So
$$
\beta_{MSD-CSC}=b\left(L_{ML-CSC}+2\right)=\beta_{ML-CSC}+2b
$$
Clearly, the $\beta$ in MSD-CSC becomes larger than that in the corresponding ML-CSC. This means MSD-CSC tends to obtain sparser solutions based on the formulation of Lasso. In addition, similar to the proof of \textbf{Theorem 1}, we can recover the dimensions of the signal which lose too much information via the identity matrix in the dictionary of MSD-CSC. All the above analysis indicates that MSD-CSC allows larger $\beta$ and lower loss of information compared to ML-CSC. MSD-CSC alleviates the contradiction between them. We can see that the identity matrix in the dictionary is an important difference between ML-CSC and MSD-CSC. Due to this, the problem (P2) could lead to a better solution for MSD-CSC compared to that of it for ML-CSC. On the other hand, the identity matrix corresponding to the dense connection in MSDNet according to \textbf{Proposition 1}. Thus, \textbf{Proposition 1} and \textbf{Theorem 1} together provide a theoretical insight into the better coding performance of dense connection. We summarize the relationship between the generalized CNNs (including ResNet and MSDNet) and the new CSC models in Table \ref{table1}.
\begin{table}
\caption{The relationship between Generalized CNN and Generalized CSC}
\label{table1}
\setlength{\tabcolsep}{2pt}
\begin{tabular}{|m{80pt}<{\centering}|m{160pt}<{\centering}|}
\hline
CNN&
CSC\\
\hline
The $i$th convolution with dilation scale $s_i$ &
The convolutional matrix $D_{i}^{s_{i}}$\\
\hline
Bias term&
The balance coefficient $\beta$ and $\lambda_{\max }\left(D^{T} D\right)$\\
\hline
ReLU  &
Soft non-negative thresholding operator $S_{\beta}^{+}(\cdot)$\\
\hline
Feed-forward algorithm&
$\Gamma^{0}=0$ in the update formula and iterate once\\
\hline
ResNet&
$\Gamma^{0}=X_{-1}$ in the update formula and iterate Eq. (3) and Eq. (5) alternately\\
\hline
Dense connection&
The identity matrix in $D_{i}^{s_{i}}$\\
\hline
\end{tabular}
\end{table}

\section{FORWARD PROPAGATION ALGORITHM}
For MSD-CSC, we just need to replace the dictionary in Eq. (1) with the corresponding one. The update formula becomes:
$$
\Gamma^{k+1}=S_{\frac{\beta}{L}}\left(\Gamma^{k}-\frac{1}{L}\left[\begin{array}{c}{I} \\ {F_{i}^{s_{i}}}\end{array}\right]\left(-\mathrm{X}+\left[\begin{array}{ll}{I} & {\left(F_{i}^{s_{i}}\right)^{T}}\end{array}\right] \Gamma^{k}\right)\right)
$$
where $I$ is an identity matrix. We can adopt ISTA and FISTA to tackle the problem (P1) \cite{Beck2009}. Besides, it is unnecessary to limit the number of iterations to be one. Iterating more than once corresponds to unfolding. The case of unfolding = 0 corresponds to MSDNet. The models with different unfolding have the same number of parameters.
According to the relationship between convolution and matrix multiplication, we can obtain the forward propagation algorithm in one layer (Fig. \ref{fig6}A) as below (Algorithm 2 and Algorithm 3).

The main difference between the two algorithms is that ISTA is only based on the last iteration, but FISTA is based on a linear combination of the last two iterations. Obviously, the main computational effort in both algorithms remains the same. We illustrate a simple architecture using MSD-CSC for a classification task (Fig. \ref{fig6}B). In this architecture, each block represents a layer in MSD-CSC. We can implement a propagating algorithm and set an unfolding number (e.g., 0, 1, 2) in each block. Note FISTA is the same as ISTA when unfolding $<$2. The feature maps in each block flow are illustrated in Fig. \ref{fig6}A. Max-pooling layers are added to downsample feature maps for memory constraints. Obviously, the number of blocks corresponds to $d$ and the number of convolution kernels corresponds to $w$ in MSDNet.

\begin{algorithm}
\caption{MSD-CSC ISTA in the $i$-th layer}
\begin{algorithmic}
\REQUIRE ~~\\ 
Signal $X$, convolution $F_{i}^{s_{i}}$, thresholding $b_{i}$, parameters $c_{i}$\\
width $w$, $\operatorname{unfolding}$
\ENSURE ~~\\  
Encoding signals $\widehat{\Gamma}_{i}$
\STATE 1. $f 1=\operatorname{conv}\left(X, F_{i}^{s_{i}}\right)$ //conv denotes convolution
\STATE 2. $\hat{\Gamma}_{i}=\operatorname{ReLU}\left(c_{i} \times \operatorname{concat}(X, f 1)+b_{i}\right)$
\STATE 3. for $k=1:\operatorname{unfolding}$ do:
\STATE 4. \quad$f1 \leftarrow \hat{\Gamma}_{i}[1:-w)+\left(F_{i}^{s_{i}}\right)^{T} \cdot \hat{\Gamma}_{i}[-w]-X$
\STATE \quad\quad\quad\quad\quad\quad\quad\quad//index $-w$ denotes the last $w$ channels
\STATE 5. \quad$f2=\operatorname{conv}\left(f 1, F_{i}^{s_{i}}\right)$
\STATE 6. \quad$f3 \leftarrow$ concat $(f1, f2)$
\STATE 7. \quad$\hat{\Gamma}_{i} \leftarrow \operatorname{ReLU}\left(\hat{\Gamma}_{i}-c_{i} \times f 3+b_{i}\right)$
\STATE 8. return $\widehat{\Gamma}_{i}$
\end{algorithmic}
\end{algorithm}

\begin{algorithm}
\caption{MSD-CSC FISTA in the $i$-th layer}
\begin{algorithmic}
\REQUIRE ~~\\ 
Signal $X$, convolution $F_{i}^{s_{i}}$, thresholding $b_{i}$, parameters $c_{i}$\\
width $w$, $\operatorname{unfolding}$, $t_{1}=1$
\ENSURE ~~\\  
Encoding signals $\widehat{\Gamma}_{i}$
\STATE 1. $f 1=\operatorname{conv}\left(X, F_{i}^{s_{i}}\right)$ //conv denotes convolution
\STATE 2. $\hat{\Gamma}_{i}=\operatorname{ReLU}\left(c_{i} \times \operatorname{concat}(X, f 1)+b_{i}\right)$
\STATE 3. $\widehat{\Gamma}_{i}^{0}=\widehat{\Gamma}_{i}^{1}$
\STATE 4. for $k=1:\operatorname{unfolding}$ do:
\STATE 5. \quad\quad$t_{k+1} \leftarrow \frac{1+\sqrt{1+4 t_{k}^{2}}}{2}$
\STATE 6. \quad\quad$Z \leftarrow \Gamma_{i}^{k}+\frac{t_{k}-1}{t_{k+1}}\left(\Gamma_{i}^{k}-\Gamma_{i}^{k-1}\right)$
\STATE 7. \quad\quad$f 1 \leftarrow Z[1:-w)+\left(F_{i}^{s_{i}}\right)^{T} \cdot Z[-w]-X$
\STATE \quad\quad\quad\quad\quad\quad\quad\quad//index $-w$ denotes the last $w$ channels
\STATE 8. \quad\quad$f2=\operatorname{conv}\left(f 1, F_{i}^{s_{i}}\right)$
\STATE 9. \quad\quad$f3 \leftarrow$ concat $(f1, f2)$
\STATE 10.\quad\quad$\Gamma_{i}^{k+1} \leftarrow \operatorname{ReLU}\left(Z-c_{k} \times f 3+b_{k}\right)$
\STATE 11. return $\Gamma_{i}^{1+\text { unfolding }}$
\end{algorithmic}
\end{algorithm}

\begin{figure*}[!t]
\centerline{\includegraphics[width=1.80\columnwidth]{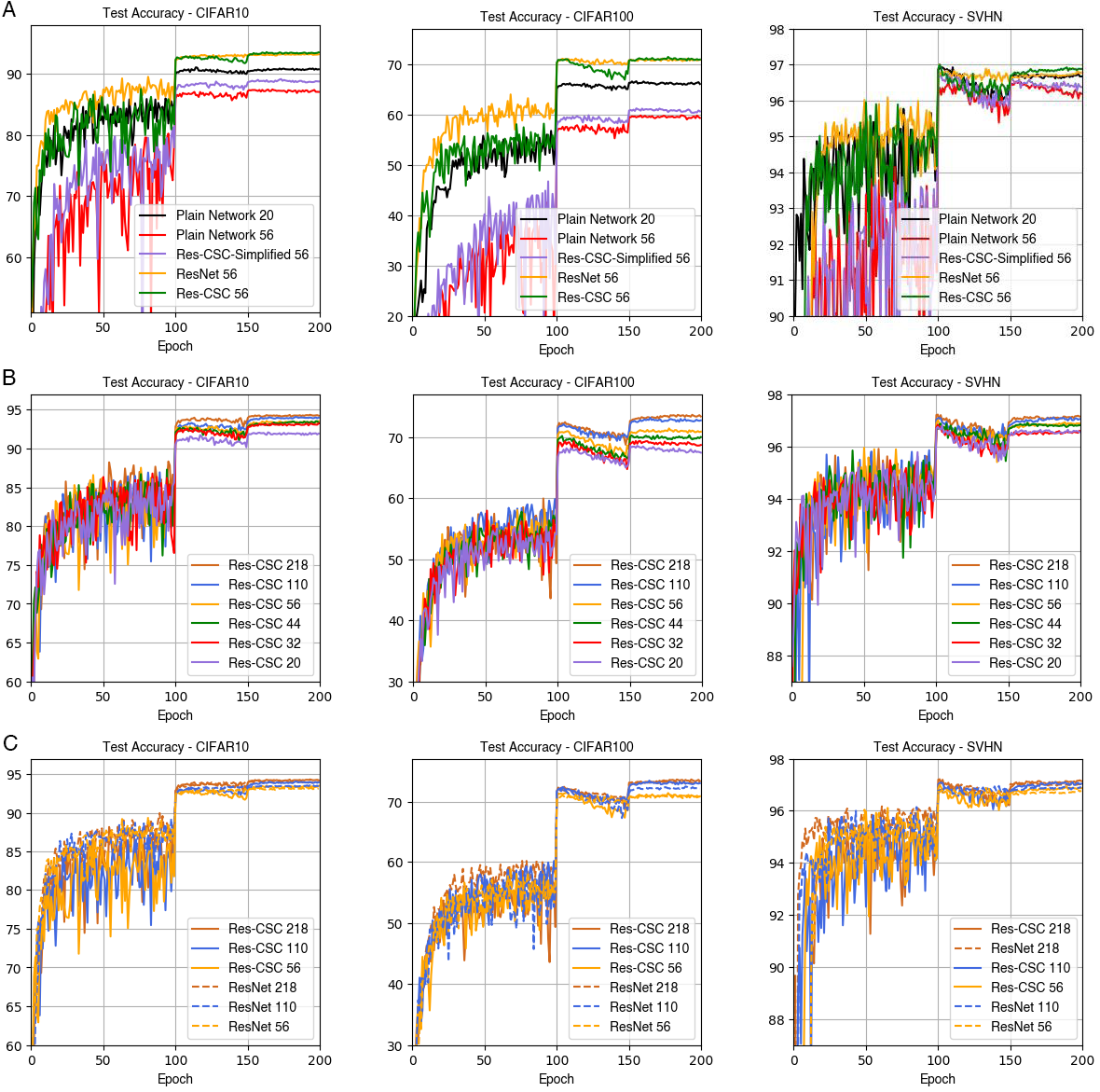}}
\caption{Experiments on Res-CSC and other methods on SVHN, CIFAR10 and CIFAR100. (A) Test accuracy versus the training epoches of the plain network Res-CSC-Simplified, ResNet and Res-CSC with 56 layers and the plain network with 20 layers for comparison. (B) Test accuracy versus the training epoches of Res-CSC with 20, 32, 44, 56, 110 and 218 layers. (C) Test accuracy versus the training epoches of Res-CSC and ResNet with 56, 110 and 218 layers.}
\label{fig7}
\end{figure*}

\section{Experiments}
In this section, we evaluate Res-CSC, MSD-CSC and related methods using three typical datasets including CIFAR10, CIFAR100 and SVHN \cite{Netzer2011}, \cite{Krizhevsky2009}. 
CIFAR10, CIFAR100 and SVHN consist of colored natural images with $32\times32$ pixels. In SVHN, the training and testing sets contain 73,257 images and 26,032 images respectively. In CIFAR10 and CIFAR100, the training and testing sets contain 50,000 and 10,000 images, which are drawn from 10 and 100 classes respectively.
\begin{table*}
\caption{The accuracy rates of a plain network, Res-CSC-Simplified, ResNet and Res-CSC on CIFAR10, CIFAR100 and SVHN}
\label{table2}
\setlength{\tabcolsep}{2pt}
\begin{threeparttable}
\begin{tabular}{|m{40pt}<{\centering}|m{40pt}<{\centering}|m{40pt}<{\centering}|m{90pt}<{\centering}|m{90pt}<{\centering}|m{90pt}<{\centering}|m{90pt}<{\centering}|}
\hline
Dataset&
Layer&
Para&
\multicolumn{4}{|c|}{Model}\\
\hline
&
&
&
Plain Network&
Res-CSC-Simplified&
ResNet&
Res-CSC\\
\hline
\multirow{5}{*}{CIFAR10}&20&0.27M&90.74\%&\textbf{91.86\%}&91.30\%&91.82\%                       \\ \cline{2-7}
                                         &32&0.46M&90.34\%&90.95\%&92.52\%&\textbf{93.31\%}      \\ \cline{2-7}
                                         &44&0.66M&89.41\%&90.19\%&\textbf{92.92\%}&92.89\%      \\ \cline{2-7}
                                         &56&0.85M&87.04\%&88.75\%&93.15\%&\textbf{93.55\%}      \\ \cline{2-7}
                                         &110&1.70M&$\times$ &$\times$ &93.41\%&\textbf{93.56\%} \\ \cline{2-7}
                                         &218&3.40M&$\times$ &$\times$ &94.02\%&\textbf{94.28\%} \\ \hline
\multirow{5}{*}{CIFAR100}&20&0.27M&66.02\%&66.33\%&67.81\%&\textbf{68.55\%}                      \\ \cline{2-7}
                                         &32&0.46M&64.31\%&65.35\%&\textbf{69.79\%}&69.35\%      \\ \cline{2-7}
                                         &44&0.66M&61.54\%&63.24\%&70.59\%&\textbf{70.68\%}      \\ \cline{2-7}
                                         &56&0.85M&59.28\%&60.58\%&71.09\%&\textbf{71.40\%}      \\ \cline{2-7}
                                         &110&1.70M&$\times$ &$\times$ &72.47\%&\textbf{73.13\%} \\ \cline{2-7}
                                         &218&3.40M&$\times$ &$\times$ &73.28\%&\textbf{73.50\%} \\ \hline
\multirow{5}{*}{SVHN}&20&0.27M&96.67\%&96.70\%&96.57\%&\textbf{96.81\%}                          \\ \cline{2-7}
                                         &32&0.46M&96.55\%&96.63\%&96.63\%&\textbf{96.83\%}      \\ \cline{2-7}
                                         &44&0.66M&96.36\%&96.51\%&96.74\%&\textbf{96.91\%}      \\ \cline{2-7}
                                         &56&0.85M&96.09\%&96.38\%&96.88\%&\textbf{97.01\%}      \\ \cline{2-7}
                                         &110&1.70M&$\times$ &$\times$ &96.96\%&\textbf{97.11\%} \\ \cline{2-7}
                                         &218&3.40M&$\times$ &$\times$ &97.05\%&\textbf{97.22\%} \\ \hline
\end{tabular}
\begin{tablenotes}
	\footnotesize
	\item[1] Top one result of the four methods in each setting is shown in boldface.
\end{tablenotes}
\end{threeparttable}
\end{table*}

\subsection{Experiments on Res-CSC}
We implement standard data augmentation (translation and horizontal flip) on CIFAR10 and CIFAR100. All the four models including a plain network, Res-CSC-Simplified, ResNet and Res-CSC are trained with the stochastic gradient descent (SGD) on a single GPU. The mini-batch size is 128 and the Nesterov momentum is set to 0.9. Each model is trained for 200 epochs. The initial learning rate is set to 0.1 and then is divided by 10 after 100 and 150 epochs.

First, we train all the four type of networks with 20 and 56 layers respectively. The plain network with 56 layers achieves lower accuracy than that with 20 layers (Fig. \ref{fig7}A and Table \ref{table2}), suggesting that the degradation phenomenon happens. Interestingly, the degradation phenomenon is alleviated to some extent with Res-CSC-Simplified though the improvement is limited. Moreover, when the number of layers comes to hundreds, Res-CSC-Simplified can not be trained normally, like the situation met in the plain network. Both Res-CSC and ResNet distinctly overcome the degradation phenomenon (Fig. \ref{fig7}B and \ref{fig7}C). These observations are consistent with our theoretical derivation. According to the results, the term $X_{-1}$ plays a more important role than the term $\mathrm{c} \cdot \mathfrak{D}^{T} \mathfrak{D} X_{-1}$ does.
\begin{figure*}[!t]
\centerline{\includegraphics[width=1.80\columnwidth]{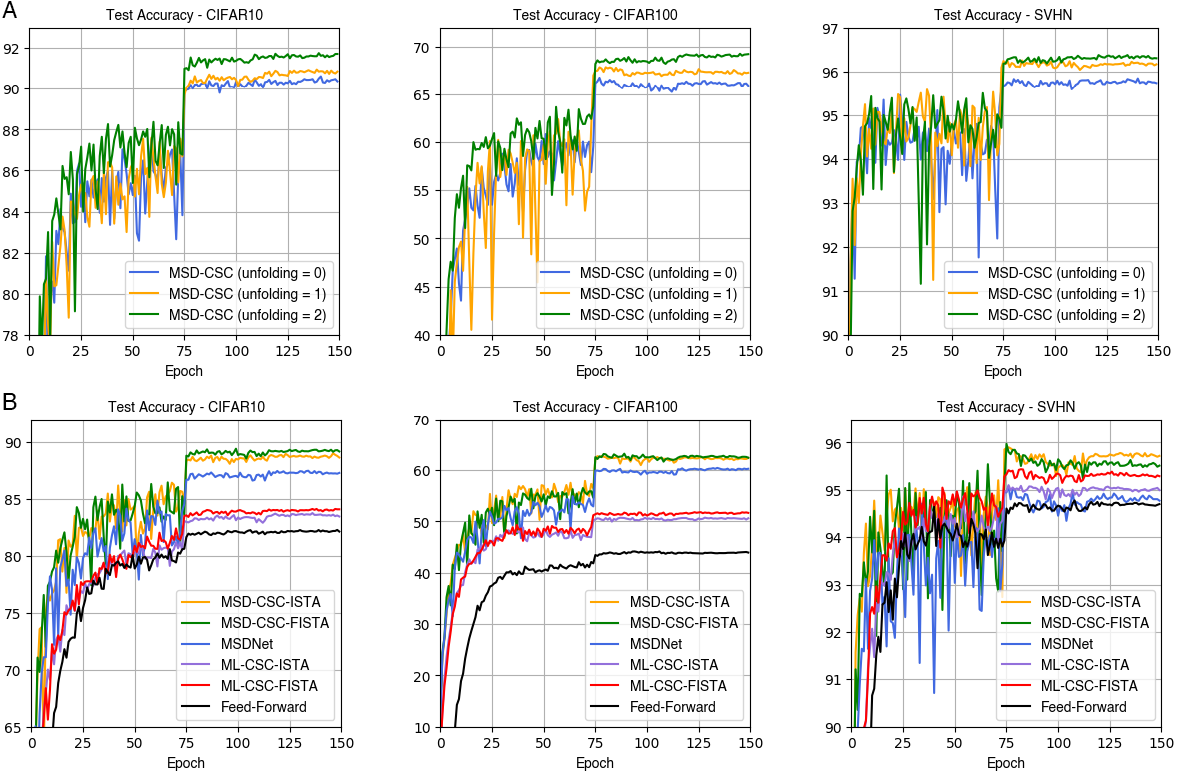}}
\caption{Experiments on MSD-CSC and other methods on SVHN, CIFAR10 and CIFAR100 with $d=12$ layers. (A) Test accuracy versus the training epoches of MSD-CSC with different unfolding$=$0, 1, 2 using FISTA. (B) Test accuracy versus the training epoches of MSD-CSC with unfolding$=$2 and other methods. }
\label{fig8}
\end{figure*}

\begin{table*}
\caption{The accuracy rates of MSD-CSC and other classic CSC models on CIFAR10, CIFAR100 and SVHN.}
\label{table3}
\setlength{\tabcolsep}{2pt}
\begin{threeparttable}
\begin{tabular}{|m{82pt}<{\centering}|m{42pt}<{\centering}|m{42pt}<{\centering}|m{42pt}<{\centering}|m{92pt}<{\centering}|m{92pt}<{\centering}|m{92pt}<{\centering}|}
\hline
Model&
Layer&
Para&
unfolding&
CIFAR10&
CIFAR100&
SVHN\\
\hline
Feed-Forward&6&4.0M&0&82.16\%&43.93\%&94.68\%\\ \hline
\multirow{2}{*}{ML-CSC-ISTA}&\multirow{2}{*}{6}&4.0M&1&83.26\%&50.23\%&94.98\%\\ \cline{3-7}
                                         &&4.0M&2&83.59\%&50.61\%&95.01\%\\ \hline
ML-CSC-FISTA&6&4.0M&2&84.06\%&51.66\%&95.29\%\\ \hline
\multirow{3}{*}{MSDNet}&6&0.1M&0&87.22\%&60.23\%&94.72\%\\ \cline{2-7}
                                         &9&0.3M&0&89.42\%&63.22\%&95.42\%\\ \cline{2-7}
                                         &12&0.6M&0&90.35\%&66.18\%&95.73\%\\ \hline
\multirow{6}{*}{MSD-CSC-ISTA}&\multirow{2}{*}{6}&\multirow{2}{*}{0.1M}&1&88.79\%&62.08\%&94.87\%\\ \cline{4-7}
                                                                                                                           &&&2&88.85\%&62.24\%&95.75\%\\ \cline{2-7}
                                         &\multirow{2}{*}{9}&\multirow{2}{*}{0.3M}&1&90.23\%&64.15\%&95.98\%\\ \cline{4-7}
                                                                                                                           &&&2&90.52\%&65.21\%&96.10\%\\ \cline{2-7}
                                         &\multirow{2}{*}{12}&\multirow{2}{*}{0.6M}&1&90.81\%&67.31\%&96.14\%\\ \cline{4-7}
                                                                                                                           &&&2&\textbf{91.76\%}&\textbf{68.14\%}&\textbf{96.36\%}\\ \hline
\multirow{3}{*}{MSD-CSC-FISTA}&6&0.1M&2&89.21\%&62.53\%&95.50\%\\ \cline{2-7}
                                                      &9&0.3M&2&90.91\%&65.22\%&95.84\%\\ \cline{2-7}
                                                      &12&0.6M&2&\textbf{91.63\%}&\textbf{69.23\%}&\textbf{96.35\%}\\ \hline

\end{tabular}
\begin{tablenotes}
	\footnotesize
	\item[1] Under the same setting, MSDNet corresponds to MSD-CSC-ISTA with unfolding $= 0$ and FISTA is the same as ISTA when unfolding $<$2. Top two cases in each dataset are shown in boldface.
\end{tablenotes}
\end{threeparttable}
\end{table*}

Next, we train six Res-CSC and ResNet models with 20, 32, 44, 56 110 and 218 layers respectively. Each Res-CSC has the same number of parameters with the corresponding ResNet. Clearly, Res-CSC achieve very competitive or even slightly better performance than ResNet in terms of accuracy (Fig. \ref{fig7} and Table \ref{table2}). The difference is due to the effect of the last term in Eq. (9). Besides, Res-CSC takes a little more time for extra convolution and transposed convolution in each layer. In short, Res-CSC is a more general white-box model for overcoming the degradation phenomenon. Its special case with $c=0$ leading to an equivalent form of ResNet. Thus, it can be an alternative to the black-box ResNet. More importantly, it leads to more theoretical understanding towards ResNet in terms of the update rule.

\subsection{Experiments on MSD-CSC}
We implement MSDNet architectures with $w=32$, $d=$6, 9, 12, $s=$1, 2, 3 and use max-pooling layers after one-third and two-thirds of the whole layers respectively (Fig. \ref{fig6}B). Before the softmax layer, average pooling is applied. We compare the results of ISTA and FISTA with different unfolding $=$0, 1, 2. In addition, we choose the traditional feed-forward network (baseline) and ML-CSC (6 layers, the kernel sizes are $4\times4$, $4\times4$, $4\times4$, $3\times3$, $3\times3$ and $3\times3$, respectively, with stride of 2 in the first three layers, and stride of 1 in the last three layers. The number of kernel in each layer are set to 32, 64, 128, 256, 512 and 512 for comparison. All these models are trained with SGD on a single GPU with momentum of 0.9. The total training epoch is set to 150. The mini-batch size is 128. The initial learning rate is set to 0.05 and is divided by 10 after 75 and 115 epochs.

First, we can clearly see that further unfolding improves the accuracy compared to MSDNet without unfolding implicitly (Fig. \ref{fig8}A and Table \ref{table3}). MSD-CSC with $d=12$ and unfolding=2 improves 1.28\% and 3.08\% on CIFAR10 and CIFAR100 respectively compared with the corresponding MSDNet. It should be emphasized that this improvement is achieved without adding any extra parameters. Second, MSD-CSC uses fewer parameters but achieves more accurate results compared to other models (Table \ref{table3} and Fig. \ref{fig8}B). That MSD-CSC has fewer number of parameters indicates it has better coding ability. This is consistent with our analysis in the Section V. Note that MSD-CSC takes more time to train though it has fewer parameters. The reason is that existing deep learning training frameworks do not support the dilation convolution and dense connection operations well since they assume that all channels of a certain feature maps are computed in the same way, and GPU convolution routines such as the cuDNN library assume that feature data is stored in a contiguous memory. Therefore, concatenate operation can be expensive in the current frameworks \cite{Pleiss2017}. Frequent concatenate and split operation are used in MSD-CSC (Fig. \ref{fig5}). This limits the application of MSD-CSC with more unfolding and layers. We believe that such an implementation issue could be addressed in the near future.

\section{Conclusion}
Inspired by the relationship between neural networks and ML-CSC, we develop the Res-CSC model to explore the layer-initializing question (LIQ). Intriguingly, ResNet can be seen as a special case of Res-CSC. Hence, we think LIQ is the key issue behind the degradation phenomenon, which has not been formally proposed before. Through evaluation on three common datasets, we find that Res-CSC reach very competitive or even slightly better performance compared to that of ResNet. Next, we introduce the MSD-CSC model to decipher the emerging MSDNet architecture via adapting the dictionaries in ML-CSC. Through the analysis of this model, we give a theoretical understanding of the dilated convolution (mixed-scale) and dense connection in MSDNet. As we know that sparse coding has more complete theory compared with neural networks. Thus, the bridge between sparse coding and neural networks will make it clear to interpret the advanced neural networks. In addition, sparse coding models can be implemented and solved with the elegant mathematical optimization algorithms such as ISTA and FISTA. Numerical experiments show that MSD-CSC performs better than ML-CSCs because of the advantage of MSDNet. Moreover, it also performs better than MSDNet because of the power of ISTA and FISTA with the unfolding trick, which achieve distinctive improvements without extra parameters.

We conclude some further thinking and potential research directions. The first one is that, as shown in this paper, Res-CSC gives an answer to LIQ. Is this answer the best? We think it is a challenge to find a universal rule for finding the best initialization since we would meet different features in different layers. Meta learning may give a potential solution through drawing lessons from MAML \cite{Finn2017}, CAML \cite{Zintgraf2018}, Reptile \cite{Nichol2018} and Meta-SGD \cite{Li2017}, which aim to learn the initializations in meta learning.

Next, we can see that dilated convolution corresponds to the dilated convolutional dictionary, and dense connection corresponds to an identity matrix in the dictionary. Can we find a better dictionary structure inspired by such a observation? And what operation does this new dictionary structure correspond to? This would help us find a new basic operation used for extracting features from data. Moreover, some architectures or operations in neural networks still lack of theoretical understanding (e.g., batch-normalization, dropout). Can we explain them in a sparse coding framework? On the one hand, we expect to find a mathematical understanding and improve the original models. On the other hand, we hope to find the key roles that these models play in the context of sparse coding.

\section*{Acknowledgment}
This work has been supported by the National Natural Science Foundation of China [11661141019, 61621003];
National Ten Thousand Talent Program for Young Top-notch Talents; National Key Research and Development Program of China [2017YFC0908405]; CAS Frontier Science Research Key Project for Top Young Scientist [QYZDB-SSW-SYS008].

\ifCLASSOPTIONcaptionsoff
  \newpage
\fi

\end{document}